\documentclass[twoside,leqno,twocolumn]{article}  
\usepackage{ltexpprt} 
\usepackage{graphicx} 
\usepackage{epsfig} 
\usepackage{multirow}

\usepackage{amssymb}
\usepackage{amsmath}
\usepackage{graphicx}
\usepackage{url}
\usepackage{algorithm}
\usepackage{algorithmic}

\begin{document}

\title{\Large Multi-task Feature Selection based Anomaly Detection}
\author{Longqi Yang\thanks{Institute of Command Information System, PLA University of Science and Technology.} 
\and 
Yibing Wang{$^*$}
\and
Zhisong Pan{$^*$}\thanks{Corresponding author. Email address: hotpzs@hotmail.com}
\and
Guyu Hu{$^*$}}
\date{}

\maketitle

\begin{abstract} \small\baselineskip=9pt 
Network anomaly detection is still a vibrant research area.
As the fast growth of network bandwidth and the tremendous traffic on the network, there arises an extremely challengeable question: How to efficiently and accurately detect the anomaly on multiple traffic?
In multi-task learning, the traffic consisting of flows at different time periods is considered as a task.
Multiple tasks at different time periods performed simultaneously to detect anomalies.
In this paper, we apply the multi-task feature selection in network anomaly detection area which provides a powerful method to gather information from multiple traffic and detect anomalies on it simultaneously. 
In particular, the multi-task feature selection includes the well-known $\ell_1$-norm based feature selection as a special case given only one task.
Moreover, we show that the multi-task feature selection is more accurate by utilizing more information simultaneously than the $\ell_1$-norm based method.
At the evaluation stage, we preprocess the raw data trace from trans-Pacific backbone link between Japan and the United States, label with anomaly communities, and generate a  248-feature dataset. 
We show empirically that the multi-task feature selection outperforms independent $\ell_1$-norm based feature selection on real traffic dataset.
\end{abstract}

\section{Introduction.}
Network anomalies typically refer to unusual and significant deviation from normal behaviors, which influence both network administrators and end users \cite{lakhina2004diagnosing}.
For the Internet Service Providers (ISPs) and network administrators, it becomes more and more important to fast and accurately classify the type of traffic and the abnormal behaviors on the backbone network.
ISPs need to monitor the constitutions of different applications so as to prioritize traffic of the QoS-sensitive application to prevent, locate harmful activities and take additional steps for other reasons, say, politics \cite{kim2008internet}.

Until now, there have been four main effective fashions on traffic classification and anomaly detection:
\begin{itemize}
\item Transport layer port number based method; 
\item Deep packet inspection;
\item Host behavior based method;
\item Traffic flow features based method.
\end{itemize}
Among them, only the fourth method belongs to the region of machine learning.

So far many states of the art methodologies are applied in the field of traffic classification and anomaly detection.
Erman, et al. \cite{erman2006traffic} used two unsupervised clustering algorithms - K-Means and DBSCAN to demonstrate how to use the cluster analysis to effectively identify groups of traffic only based on the transport layer statistics.
Brauckhoff et al. \cite{brauckhoff2009applying} applied PCA for traffic anomaly detection overcoming the sensitive to its parameter setting by a slight modification.
Kim, et al. \cite{kim2008internet} conducted an evaluation of three traffic classification approaches: port-based, host-behavior-based and flow-features-based. By comparing seven commonly used traffic-flow-features-based methods with the other two kinds of traffic classification methods, they found that Support Vector Machine (SVM) algorithm achieved the highest accuracy on every trace and application.  
$\ell_1$-norm minimization Extreme Learning Machine (ELM) \cite{yibingelm} has reached a good performance on anomaly detection.

Most of the mentioned methods above were applied in the circumstance of single traffic, and difficultly to meet the demand of complex multiple traffic cases.
Learning from multiple tasks simultaneously has been widely applied \cite{zhou2011multi,zhou2012modeling,zhou2013modeling,zhoupatient,zhou2013feafiner,obozinski2006multi,argyriou2008convex,zhang2008flexible,bi2008improved,torralba2004sharing,ando2006applying} and shown to significantly improve the performance relative to learning each task independently \cite{argyriou2006multi}. 
Multi-task learning has been successfully applied in various areas including: handwriting character recongnition \cite{obozinski2006multi}, predicting disease progression \cite{zhou2011multi}, modeling disease progression \cite{zhou2012modeling,zhou2013modeling}, patient risk prediction \cite{zhoupatient}, biomarker identification \cite{zhou2013feafiner}, conjoint analysis \cite{argyriou2008convex,zhang2008flexible}, medical diagnosis \cite{bi2008improved}, computer vision \cite{torralba2004sharing}, natural language processing \cite{ando2006applying} and text classification \cite{zhang2008flexible}.

In multi-task learning, the traffic consisting of flows at each time periods is considered as a task. 
These multiple tasks at different time periods performed are learnt simultaneously by extracting and utilizing appropriate shared information across tasks.
The useful features learnt from multiple tasks can significantly improve the performance on anomaly detection.

In this paper, our main contributions can be summarized below:
\begin{itemize}
\item Applying multi-task feature selection on network anomaly detection area.
\item Employing an effective preprocessing and feature extraction step to deal with the raw network data trace. 
\item Generating anomaly detection datasets with ground truth, which can be used comprehensively.
\end{itemize}

To the best of our knowledge, it is the first time that multi-task feature selection method is applied in network anomaly detection area.
Although the application is still in its infancy, we believe that it has great potential to enhance the field.

The rest of paper is organized as follows: Section 2 presents the idea of $\ell_1$-norm based feature selection, and discusses the multi-task feature selection extension. Section 3 describes how to convert from raw network data trace into features and process of labeling . Section 4 presents systematic evaluation of different methods for anomaly detection. Conclusions come to in Section 5.

\section{Multi-task Feature Selection.}
In this section, we review the work of $\ell_1$ regularization - Lasso \cite{tibshirani1996regression} and the extending of the well-known $\ell_{1}$ regularization for single-task to multi-task setting. 

Formally, we assume there are $L$ models to learn and the training set consists of  
$\{{(a_i^j,y_i^j)}\}_{i=1}^{m_j},  j=1,2,\ldots,L$, 
where $a_i^j\in {\mathbb{R}^n}$ denotes the training sample for the $jth$ task, $y_i^j$ denotes the corresponding output, $m_j$ is the number of training sample for $jth$ task, 
and $m=\sum_{j=1}^{k}m_j$ is the total number of training samples.
n denotes the number of features. 
Let $A_j = [a_1^{j},\ldots,a_{m_j}^j]^{T}\in{\mathbb{R}^{m_j\times{n}}}$ denote the data matrix for the $jth$ task, $A=[A_1^T,\ldots,A_L^T]\in{\mathbb{R}^{m\times{n}}}$, $y_j=[y_1^j,\ldots,y_{m_j}^j]^T\in{\mathbb{R}^{m_j}}$, 
and $y=[y_1^T,\ldots,y_k^T]^T\in{\mathbb{R}^m}$.

\subsection{Single task case.}
For single-task case, the well-known $\ell_{1}$-norm regularization implemented by Lasso method produces a few non-zero coefficients that gives a more accurate and more easily interpretable model. 
Learning a model independently through minimizing empirical risk with $\ell_{1}$ regularization:
 
 \begin{equation}
 \min_{w_i} \, \frac{1}{2}\Arrowvert {y_{i}-A_{i}w_{i}} \Arrowvert_2^2+\lambda\Arrowvert{w_i}\Arrowvert_1
 \end{equation}
 
 Solving these problems individually is equivalent to the solving of the summing problem:
  \begin{equation}
 \min_{W} \, \frac{1}{2}\sum_{i=1}^{L}\Arrowvert {y_{i}-A_{i}w_{i}} \Arrowvert_2^2+\lambda\sum_{i=1}^{L}\Arrowvert{w_i}\Arrowvert_1
 \end{equation}
 
 It yields to a sparse $w_i$ for each task with distinct sparsity pattern by solving this optimization problem. 
 $w^i$ denotes the column of $W$ which is the weight vector of $ith$ feature for all $L$ tasks, and $w_j$ denotes the $jth$ row of $W$, where $W$ stands for the weight matrix.

 \subsection{Extending from single-task case to multi-task case.}
In network traffic anomaly detection area, the data collected range from a certain network at different time to different backbone network sources forms multi-task cases. 
 Our goal is to learn from a group of tasks by considering the relations between them resulting in a shared pattern of sparsity. 
 Obzinski et al. \cite{obozinski2006multi} proposed a joint regularization of the parameters which selects features across a group of tasks. 
 The joint regularization blocks the weight vector of a certain feature across all tasks with the $\ell_{2}$-norm. 
 To penalize the $\ell_{1}$-norm of these blocks which result in global feature selection.
The optimization problem is to minimize the following $\ell_{2,1}$-norm regularization:
 
\begin{equation}\label{l21}
 \min_{W} \,  \frac{1}{2}\Arrowvert {y-AW} \Arrowvert_2^2+\lambda\Arrowvert{W}\Arrowvert_{2,1}
\end{equation}

where $\Arrowvert{W}\Arrowvert_{2,1}=\sum_{i=1}^{n}\Arrowvert{w^i}\Arrowvert_2$.

The nonsmooth \ref{l21} could be solved by accelerated gradient methods (AGM). The AGM is the optimal among first order methods, which has the convergence speed of $O(\frac{1}{k^2})$. 
%
%
%
%
%

 \section{Data Preprocessing.}
 Dataset is generated broadly including three main steps: preprocess the raw traffic trace, the feature extraction step and the labeling step. The process is briefly described in figure 1.
 
 \begin{figure}[h]
 \centering\includegraphics[width=8cm]{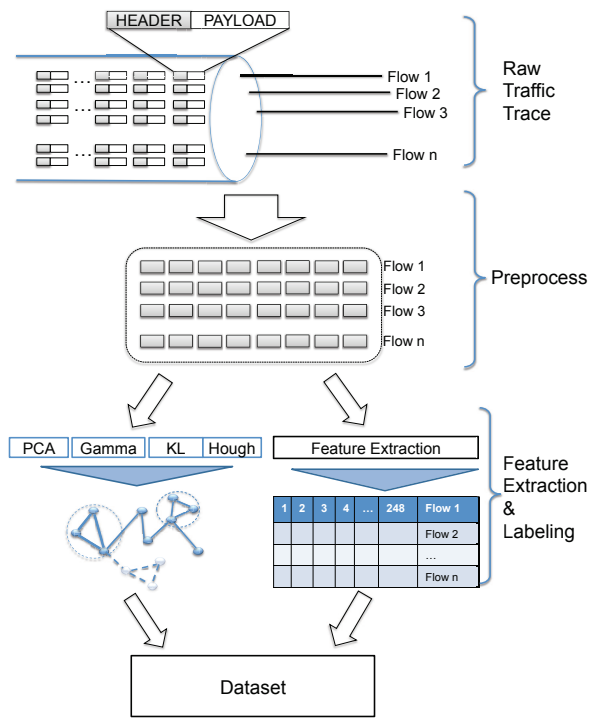} 
 \caption{From raw data traces to labeled dataset.}
 \end{figure}

  \begin{table}[b]
 \caption{traffic trace statistics}
 \footnotesize\centering
  \begin{tabular}{r|p{3cm}}
 \hline
 Number of flows & 731,362 \\
 Seconds of duration	& 899.20 \\
Number of packets	& 28,944,127 \\
Number of IPv4 addresses & 420,421 \\
Number of IPv6 addresses & 2,893 \\
Average Rate & 194.17Mbps \\
Total Captured Size	& 1588.48MB \\
Total Flow Size	& 20809.67MB \\
 \hline
 \end{tabular}
 \end{table}
 
 \subsection{Raw Traffic Traces.} 
 
 In the traffic classification and anomaly detection research field, owing to the privacy and legal concern, it is lack of the benchmark datasets that could be used to verify the feature selection methods and classifiers.
The WIDE project \cite{cho2000traffic} maintains a data trace repository. Traffic traces were captured from a trans-Pacific backbone link between Japan and the United States.  A 15-minute traffic trace is chosen as the base evaluation dataset that started from 14:00 and ended at 14:15 on January 9th, 2011. The statistics of traffic trace could be found in table1.

 \subsection{Feature Extraction.}
 Moore et al. \cite{moore2005discriminators} proposed comprehensive discriminators to characterize the flow which provided 248 features. 
The more detailed the features could be, the more proper model would achieve better classification performance.
 The features are calculated through complete TCP flows which derived from packet header. 
 A TCP flow is defined as one or more packets traveling between two computer addresses using TCP protocol. 
 Every packet contains 5-tuple which is made up of source IP address, destination IP address, source port number, destination port number and the protocol in use. 
 Netdude is used to create a set of complete TCP flows on account of packet loss.
A variety of features are selected to characterize the TCP flow includes packet length, inter-packet timings and information about transport protocol(TCP): such as SYN and ACK counts. 
Tcptrace can deal with some packet statistics which need to estimate round-trip time, size of TCP segments.
However, for the simpler statistics such as counting packets and packet header size, calculation is more efficient. 
The 248 features could be found in table2. 
 
 \begin{table}[htbp!]
 \centering
 \caption{248 features derived from complete tcp flows}
 \footnotesize
 
 \begin{tabular}{r|p{6cm}}
 \hline
 Sequence & Content \\
\hline
1 &   Port number at server \\
2 &   Port number at client   \\
3 &   Minimum packet inter-arrival time for all packets of the flow \\
4 &	First quartile inter-arrival time \\
5 &	Median inter-arrival time \\
6 & 	Mean inter-arrival time \\
7 &	Third quartile packet inter-arrival time \\
\ldots & \ldots \\
25 & 	First quartile of control bytes in packet \\
26 &	Median of control bytes in packet \\
27 &	Mean of control bytes in packet \\
\ldots & \ldots \\
239 & FFT of packet IAT (Frequency 1) \\
\ldots & \ldots \\
248 &  FFT of packet IAT (Frequency 10) \\

 \hline
 \end{tabular}
\end{table}

\subsection{Labeling.}
Due to the lack of ground truth, labeling the traffic set is difficult.
The method \cite{fontugne2010mawilab} to label the dataset which includes four main steps:

\begin{enumerate}
\item Four anomaly detectors analyze the traffic and report alarms
\item Uncover the similarities among the reported alarms by using similarity estimator, then group similar alarms into communities. 
\item Combiner classifies each community according to overall output of all detectors.
\item Labeling the datasets.
\end{enumerate}
 
\subsubsection{Analyze the traffic and report alarms.} 
The four unsupervised anomaly detectors report traffic at different granularities.

\begin{enumerate}
\item PCA: PCA-based detector reports the source IP address of identified anomalous traffic by using random projection techniques. 
\item Sketching and multi-resolution gamma modeling based method \cite{dewaele2007extracting}: This method reports the anomalous traffic that is distant from reference. The traffic is split into sketches, then modeled using Gamma distribution.
\item Hough transform based method \cite{fontugne2011hough}: an useful technique for identification of a specific shape in a picture.  
\item Kullback-Leibler divergence based method: detected the prominent changes in traffic. The alarms reported by this anomaly detector are association rules, namely 4-tuples (source and destination IP addresses, source and destination port numbers).
\end{enumerate}

\subsubsection{Extract the traffic with similar alarms and group into communities.}

 The traffic extractor selects the traffic described by each alarm. 
 Using both packet and flow as granularities could avoid missing similar alarms. 
 \begin{figure}[h]
 \centering\includegraphics[width=7cm]{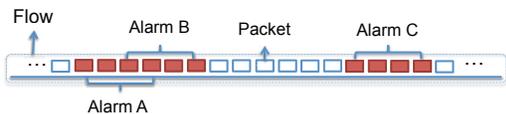} 
 \caption{AlarmC has no common packets shared with AlarmA and AlarmB under the packet granularity. 
 Whereas the three alarms have similarities by using flow granularity.
}
 \end{figure}
 
 The graph generator builds an undirected graph by using the retrieved traffic. 
 In this similarity graph, nodes stand for alarms and an edge between two nodes if their associated traffic intersects.  
The identical alarms in the graph are a set of the strong connected nodes, which is called community. 
After identifying the community and combining the communities, each community is classified to either accepted or rejected by measuring the distance to the reference communities in the low-dimensional space.

\subsubsection{Labeling the datasets.}
 To label the analyzed traffic, we define a simple traffic taxonomy with two labels: anomalous and normal. 
 Where the anomalous label stands for the traffic accepted.
 And this traffic should be detected as anomalous by any efficient anomaly detectors. 
 The label normal consists of two cases. 
 One is the traffic rejected which has a relative distance to the reference points. 
 The other one is none of the detectors identified the traffic.

 \section{Experiment}
 In the experiment we study the empirical performance of feature selection algorithms and classification methods in datasets which processed in Section 3.
In this section, in order to show the effectiveness of the multi-task feature selection, we compare it with single-task feature selection method - Lasso.
 
 \subsection{Dataset}
 After preprocessing the raw traffic trace, we formed a dataset consisted of flow discriminators. 
 Entire data flow is fully characterized in a row. 
 The dataset consisting of 235000 flows which is splited into 10 tasks chronologically. 
We select 50\% samples for training and 50\% for testing. 
The statistics of the dataset and samples could be found in table3. 
 
 \begin{table}[t]
 \caption{The sample size of the tasks.}
 \footnotesize
\centering
\begin{tabular}{c|c|cc|cc} 
\hline 
\multicolumn{1}{c|}{\multirow {2}{*}{Task\#}} & \multicolumn{1}{c|}{\multirow {2}{*}{\#of flow}}& \multicolumn{2}{c|}{Training} & \multicolumn{2}{c}{Testing}   \\
\cline{3-4} \cline{5-6} \multicolumn{1}{c|}{}& \multicolumn{1}{|c|}{} & \multicolumn{1}{|c} {+} & - & \multicolumn{1}{c}{+} & - \\
\hline 
 01 & 23503 & 10845 & 907 & 10787 & 964 \\
 02 & 23504 & 10007 & 1745 & 9928 & 1824 \\
 03 & 23503 & 10082 & 1670 & 10061 & 1690 \\
 04 & 23503 & 10228 & 1524 & 10203 & 1548 \\
 05 & 23503 & 10268 & 1484 & 10174 & 1577 \\
 06 & 23503 & 10134 & 1618 & 10085 & 1666 \\
 07 & 23503 & 10092 & 1660 & 10175 & 1576 \\
 08 & 23503 & 9854 &  1898 & 9833 & 1918 \\
 09 & 23503 & 10010 & 1652 & 10121 & 1630 \\
 
 10 & 23503 & 10071 & 1681 & 10092 & 1659\\
 \hline
 \end{tabular}
 \end{table}
 
 \subsection{Experiment Setup}
 In the experiments the implementation of multi-task feature selection and Lasso from the MALSAR package \cite{zhou2012malsar} and SLEP package \cite{liu2009slep}, respectively. SVM is chosen to be the classifier and implemented by Matlab version of libsvm \cite{chang2011libsvm}. 
 For multi-task case, the multi-task feature selection method is applied to ten tasks simultaneously by tuning the parameter of $\lambda$, which selected 5, 12 and 24 features respectively. For single-task case, we choose top $t$ features by turning the parameter $\lambda$, where varying t from 5, 12 to 24. 
After the feature selection, we build classification model using SVM and evaluate the model on the testing data.
For SVM, 5-fold cross validation is used to estimate the best parameter $c$ and $g$.

 \subsection{Result}
To measure the performance of multi-task feature selection method, we use overall accuracy as metric.
In figure 3 and figure 5, we compare the overall accuracy of SVM by selecting top-5 features and top-20 features.
The multi-task feature selection achieve the highest overall accuracy in 8 of all 10 tasks.
In figure 4, we use top-12 features to measure the performance.
With the similar result, multi-task feature selection method runs ahead in 9 of all 10 tasks.
We can observe that the performance improves slightly with increasing of the number of selected feature and the classifier works well using only 5 features.
Since the class distribution of the data sets is not balanced, we also focus on the Area Under Curve (AUC) metric.
From table 4, We observe that multi-task feature selection method achieve better performance than Lasso on both top-5 features case and top-12 features case.

\begin{table}[h]
\caption{SVM performance in terms of AUC.}
\footnotesize
\centering
\begin{tabular}{c|c|c|c|c} 
\hline 
\multicolumn{1}{c|}{\multirow {2}{*}{Task\#}}& \multicolumn{2}{|c}{Features=5} & \multicolumn{2}{|c}{Features=12}   \\
\cline{2-3} \cline{4-5} \multicolumn{1}{c|}{}& \multicolumn{1}{|c|}{Lasso} & Multi-task & \multicolumn{1}{|c|}{Lasso} & Multi-task \\
\hline 

01            & 0.8651 & \textbf{0.9190} & \textbf{0.9184} & 0.9141 \\
02           & \textbf{0.9424} & 0.9365 & \textbf{0.9401} & 0.9389  \\
03 &       0.9451& \textbf{0.9549} & \textbf{0.9600} & 0.9596 \\
04 & 0.9426 & \textbf{0.9481} & 0.9366 & \textbf{0.9396} \\
05 & 0.9401 & \textbf{0.9454} & 0.9318 & \textbf{0.9333} \\
06 & 0.9294 & \textbf{0.9294} & 0.9458 & \textbf{0.9641}\\
07 & 0.9062 & \textbf{0.9221} &0.9402 & \textbf{0.9437} \\
08 & 0.9387 & \textbf{0.9387} & 0.9417 & \textbf{0.9709} \\
09 & 0.9378 & \textbf{0.9414} & 0.9302 & \textbf{0.9606} \\
10 & \textbf{0.9270} & 0.9244 & 0.9456 & \textbf{0.9618} \\
\hline
\end{tabular} 
\end{table}

 \section{Conclusion}
In this paper, we apply a multi-task feature selection method on network anomaly detection field.
The method selects effective features from multiple tasks simultaneously which significantly improves the performance on detecting anomalies.
We employ an effective preprocessing and feature extraction step to deal with the real data trace from backbone network. Then generate an anomaly detection dataset with ground truth.
The datasets generated with those features helped to evaluate the multi-task feature selection method, which outperforms the Lasso in most cases.
We plan to extend the scale of application by evaluating the multi-task feature selection method on different backbone network simultaneously, which could be a network-wide anomaly detection.

  \begin{figure}[h]
 \centering\includegraphics[width=8cm]{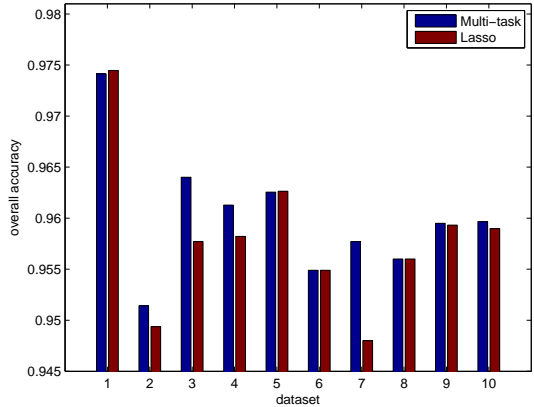} 
 \caption{Overall accuracy on top 5 features}
 \end{figure}
 
 \begin{figure}[h]
\centering \includegraphics[width=8cm]{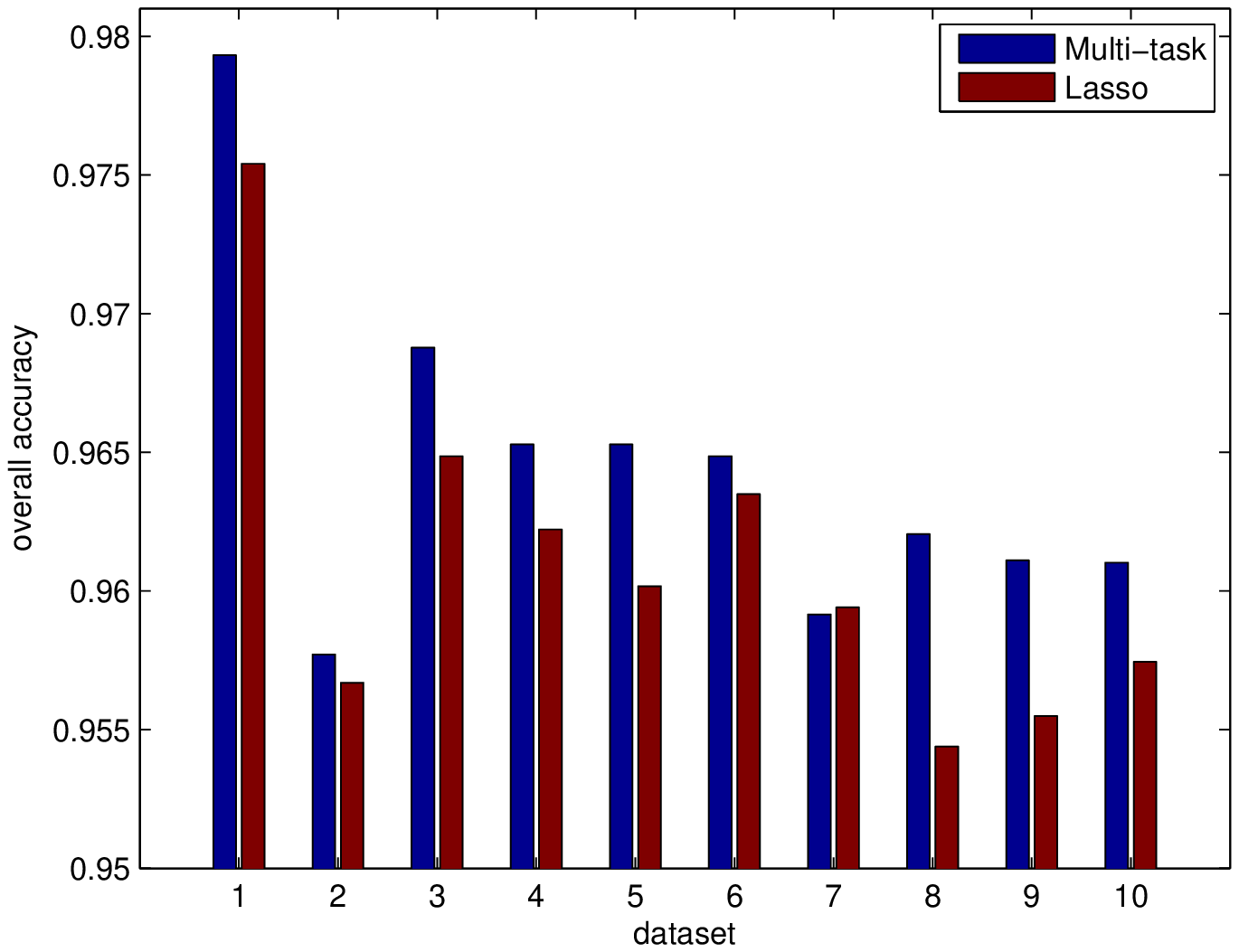}
\caption{Overall accuracy on top 12 features}
 \end{figure}

\begin{figure}[h]
\centering \includegraphics[width=8cm]{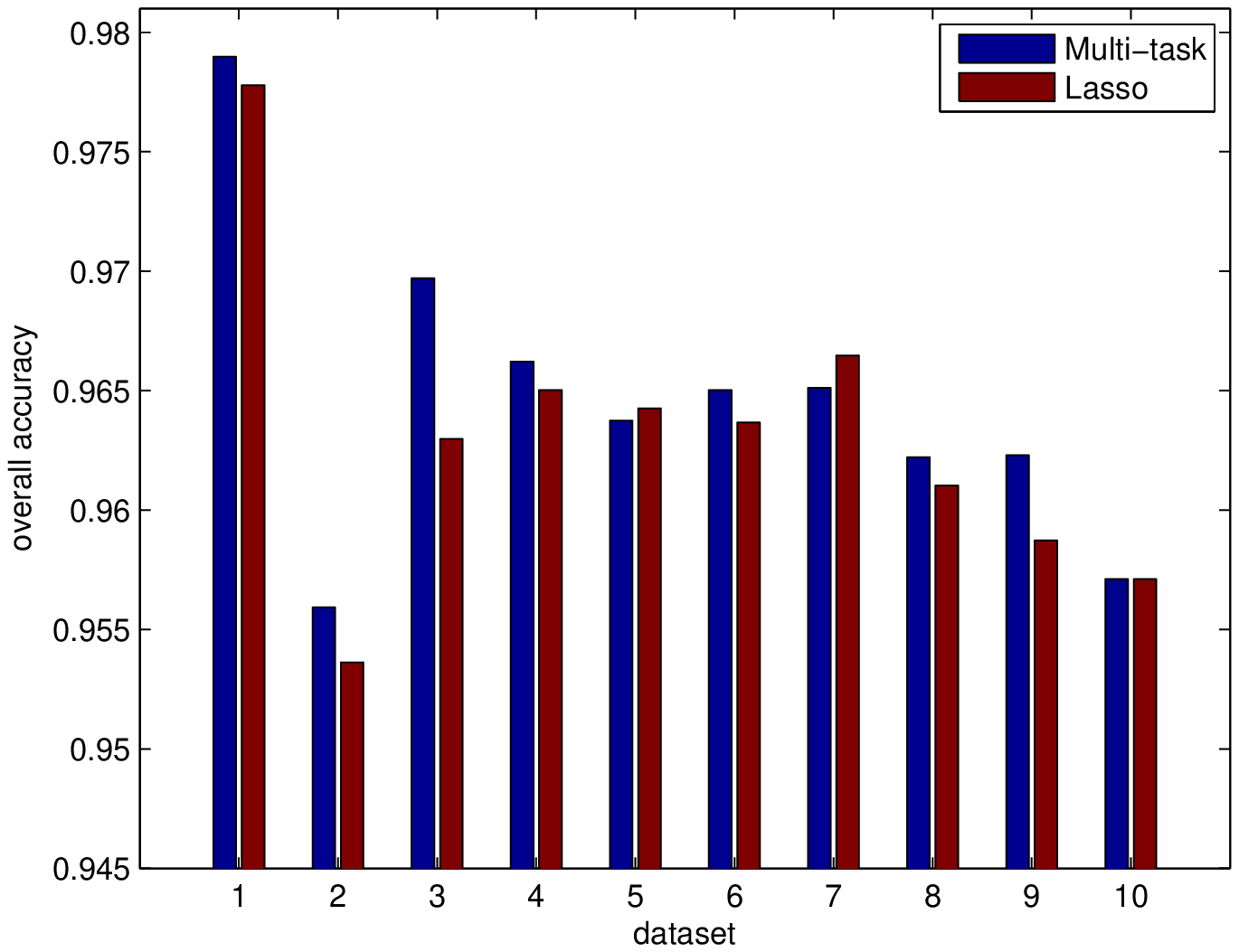}
\caption{Overall accuracy on top 24 features}
 \end{figure}

\bibliographystyle{unsrt}
\bibliography{./multi-task_feature_selection_based_anomaly_detection}

\end{document}